\documentclass{article}
\pdfoutput=1

\usepackage[preprint,nonatbib]{nips_2018}

\usepackage[utf8]{inputenc} % allow utf-8 input
\usepackage[T1]{fontenc}    % use 8-bit T1 fonts
\usepackage{hyperref}       % hyperlinks
\usepackage{url}            % simple URL typesetting
\usepackage{booktabs}       % professional-quality tables
\usepackage{amsfonts}       % blackboard math symbols
\usepackage{nicefrac}       % compact symbols for 1/2, etc.
\usepackage{microtype}      % microtypography

\usepackage{graphicx}
\usepackage{subfigure}

\usepackage{bm}
\usepackage{amsmath}
\usepackage{amsfonts}
\usepackage{xcolor}

\title{An Information-Theoretic Optimality Principle for Deep Reinforcement Learning}

\author{
  Felix Leibfried, Jordi Grau-Moya, Haitham Bou-Ammar \\
  PROWLER.io\\
  Cambridge, UK \\
  \texttt{\{felix,jordi,haitham\}@prowler.io}
}

\begin{document}

\maketitle

\begin{abstract}
  We methodologically address the problem of Q-value overestimation in deep reinforcement learning to handle high-dimensional state spaces efficiently. By adapting concepts from information theory, we introduce an intrinsic penalty signal encouraging reduced Q-value estimates. The resultant algorithm encompasses a wide range of learning outcomes containing deep Q-networks as a special case. Different learning outcomes can be demonstrated by tuning a Lagrange multiplier accordingly. We furthermore propose a novel scheduling scheme for this Lagrange multiplier to ensure efficient and robust learning. In experiments on Atari, our algorithm outperforms other algorithms (e.g. deep and double deep Q-networks) in terms of both game-play performance and sample complexity. These results remain valid under the recently proposed dueling architecture.
\end{abstract}

\section{Introduction}
\label{sec:introduction}

Reinforcement learning~\cite{Sutton1998} (RL) is a discipline of artificial intelligence seeking to find optimal behavioral policies that enable agents to collect maximal reward while interacting with the environment. A popular RL algorithm is Q-learning \cite{Watkins1989} that operates by estimating expected cumulative rewards (Q-values). Although successful in numerous applications~\cite{Busoniu2010}, standard Q-learning suffers from two drawbacks. First, due to its tabular nature in representing Q-values, it is not readily applicable to high-dimensional environments with large state and/or action spaces. Second, it initially overestimates Q-values, introducing a bias at early stages of training \cite{Fox2016}. This bias has to be ``unlearned'' as training proceeds, thus decreasing sample efficiency.

To address the first problem, Q-learning has been extended to high-dimensional environments by using parametric function approximators instead of Q-tables~\cite{Busoniu2010}. One particularly appealing class of approximators are deep neural networks that learn ``complex'' relationships between high-dimensional inputs (e.g. images) and low-level actions. Building on this idea, deep Q-networks (DQNs) \cite{Mnih2015} were proposed, attaining state-of-the-art results in large-scale domains, e.g. the Arcade Learning Environment for Atari games \cite{Bellemare2013}. Though successful, DQNs fail to address the overestimation problem, and are therefore rather sample-inefficient~\cite{vanHasselt2016}.

One way of addressing Q-value overestimation is to introduce an intrinsic penalty signal in addition to instantaneous rewards. The intrinsic penalty affects the learned Q-values, eventually leading to lower estimates. Information theory provides a principled method to formalize such a penalty by interpreting the agent as an information-theoretic channel with limited transmission rate~\cite{Sims2010,Ortega2013}. Specifically, the state of the environment is interpreted as channel input, the action as channel output and the agent's reward as quality of information transmission~\cite{Genewein2015}. Interestingly, in the RL setting, limits in transmission rate reflect limits in ``information resources'' the agent can spend to deviate from a given reference policy. The instantaneous deviation between the agent's current policy and such a reference policy directly results in an intrinsic penalty to be subtracted from the reward. Information-theoretic RL approaches \cite{Azar2012,Rawlik2012,Fox2016} have been designed for the tabular setting but do not readily apply to high-dimensional environments that require parametric function approximators.

Since we are interested in improving sample complexity of RL in high-dimensional state spaces, we contribute by adapting information-theoretic concepts to phrase a novel optimization objective for learning Q-values with deep parametric function approximators. The resultant algorithm encompasses a wide range of learning outcomes that can be demonstrated by tuning a Lagrange multiplier. We show that DQNs arise as a special case of our proposed approach. We further contribute by introducing a dynamic scheduling scheme for adapting the magnitude of intrinsic penalization based on temporal Bellman error evolution. This allows us to outperform DQN and other methods, such as double DQN \cite{vanHasselt2016} and soft Q-learning~\cite{Schulman2017}, by large margins in terms of game score and sample complexity in the Atari domain. At the same time, our approach leads to decreased Q-value estimates, confirming our hypothesis that overestimation leads to poor performance in practice.
Finally, we show further performance increase by adopting the dueling architecture from \cite{Wang2016}.
In short, our contributions are: 

\begin{enumerate}
\item applying information-theoretic concepts to large state spaces with function approximators; 
\item proposing a novel information-theoretically inspired optimization objective for deep RL;
\item demonstrating a wide range of learning outcomes for deep RL with DQN as a special case; 
\item and outperforming DQN, double DQN, and soft Q-learning in the Atari domain. 
\end{enumerate}

\section{Reinforcement Learning}
\label{sec:overestimations}

In RL, an agent, being in a state $\bm{s} \in \mathcal{S}$, chooses an action $\bm{a} \in \mathcal{A}$ sampled from a behavioral policy $\bm{a} \sim \pi_{\text{behave}}(\bm{a}|\bm{s})$, where $\pi_{\text{behave}}: \mathcal{S}\times \mathcal{A} \rightarrow [0,1]$. Resulting from this choice is a transition to a successor state $\bm{s}^{\prime}\sim \mathcal{P}\left(\bm{s}^{\prime}|\bm{s},\bm{a}\right)$, where $\mathcal{P}:\mathcal{S}\times \mathcal{A}\times \mathcal{S} \rightarrow [0,1]$ is the unknown state transition model, and a reward $r = \mathcal{R}(\bm{s},\bm{a})$ that quantifies instantaneous performance. After subsequent interactions with the environment, the goal of the agent is to optimize for $\pi^{\star}_{\text{behave}}$ that maximizes the expected cumulative return $\mathbb{E}_{\pi_{\text{behave}},\mathcal{P}}\left[\sum_{t=0}^{\infty} \gamma^{t}r_{t}\right]$, with $t$ denoting time and $\gamma \in (0,1)$ the discount factor. 

Clearly, to learn an optimal behavioral policy, the agent has to reason about long term consequences of instantaneous actions. Q-learning, a famous RL algorithm, estimates these effects using state-action value pairs (Q-values) to quantify the performance of the policy. In Q-learning, updates are conducted online after each interaction $(\bm{s},\bm{a},r,\bm{s'})$ with the environment using
\begin{equation}
\label{Eq:QLearning}
Q\left(\bm{s},\bm{a}\right) \leftarrow Q(\bm{s},\bm{a}) + \alpha \left(r + \gamma \max_{\bm{a}^{\prime}} Q(\bm{s}^{\prime},\bm{a}^{\prime}) - Q\left(\bm{s},\bm{a}\right) \right),
\end{equation}
with $\alpha >0$ being a learning rate. Equation~\eqref{Eq:QLearning} assumes an old value, i.e. the prediction $Q(\bm{s},\bm{a})$, and corrects for its estimate based on new information, i.e. the target $r + \gamma \max_{\bm{a}^{\prime}} Q(\bm{s}^{\prime},\bm{a}^{\prime})$.

\paragraph{Optimistic Overestimation:}

Upon careful investigation of Equation~\eqref{Eq:QLearning}, one comes to recognize that Q-learning updates introduce a bias to the learning process caused by an overestimation of the optimal cumulative rewards~\cite{vanHasselt2010,Azar2011,Lee2012,Bellemare2016,Fox2016}. Specifically, the usage of the maximum operator assumes that current guesses for Q-values reflect optimal cumulative rewards. Of course, this assumption is violated, especially early in the learning process, when a relatively small number of updates has been performed. Due to the correlative effect of ``bad'' estimations between different state-action pairs, these mistakes tend to propagate rapidly through the Q-table and have to be unlearned in the course of further training. Though such an optimistic bias is eventually unlearned, the convergence speed (in terms of environmental interactions, i.e. sample complexity) of Q-learning is highly dependent on the quality of the initial Q-values.

The problem of optimistic overestimation only worsens in large state spaces, such as images in Atari. As mentioned earlier, high-dimensional representations are handled by generalizing tabular Q-learning to use parametric function approximators, e.g. deep neural networks~\cite{Mnih2015}. Learning then commences by fitting weights of the approximators using stochastic gradients to minimize
\begin{equation}
\label{Eq:dqlearning}
\mathbb{E}_{\bm{s},\bm{a},r,\bm{s}^{\prime}} \left[ \left( r + \gamma \max_{\bm{a}^{\prime}} Q_{\bm{\theta}^{-}}(\bm{s}^{\prime},\bm{a}^{\prime}) - Q_{\bm{\theta}}(\bm{s},\bm{a}) \right)^2 \right].
\end{equation}
Here, the expectation $\mathbb{E}$ refers to samples drawn from a replay memory storing state transitions \cite{Lin1993}, and $Q_{\bm{\theta}^{-}}(\bm{s}^{\prime},\bm{a}^{\prime})$ denotes a DQN at an earlier stage of training. The minimization objective in Equation~\eqref{Eq:dqlearning} resembles similarities to that used in the tabular setting. Again, old value estimates are updated based on new information, while introducing the $\max$-operator bias. Although DQNs generalize well over a wide range of input states, they are ``unaware'' of the aforementioned overestimation problem \cite{Thrun1993}. However, when compared with the tabular setting, this problem is even more severe due to the lack of any convergence guarantees to optimal Q-values when using parametric approximators, and the inability to explore the whole state-action space. Hence, the number of environmental interactions needed to unlearn the optimistic bias can become prohibitively expensive.

\section{Addressing Optimistic Overestimation}
\label{sec:info_core}

A potential solution to optimistic overestimation in Q-learning is to add an intrinsic penalty to instantaneous rewards, thus reducing Q-value estimates. A principled way to introduce such a penalty is provided by the framework of information theory for decision-making. The rationale is to interpret the agent as an information-theoretic channel with limited transmission rate \cite{Sims2010,Tishby2011,Ortega2013,Genewein2015}. The environmental state $\bm{s}$ is considered as channel input, the agent's action $\bm{a}$ as channel output and the quality of information transmission is expressed by some reward or utility function $U(\bm{s}, \bm{a})$. According to Shannon's noisy-channel coding theorem \cite{Shannon1948}, the transmission rate is upper-bounded by the average Kullback-Leibler (KL) divergence between the behavioral policy $\pi_{\text{behave}}$ and any arbitrary reference policy with support in $\mathcal{A}$ \cite{Csiszar1984,Tishby1999}. In the following, the reference policy is denoted as prior policy $\pi_{\text{prior}}$. The KL-divergence, therefore, plays the role of a limited resource and may not exceed a maximum $K>0$, such that $\text{KL}\left(\pi_{\text{behave}}||\pi_{\text{prior}}\right) \leq K$.

The intuition behind the information-theoretic viewpoint is that the channel aims to map input $\bm{s}$ to output $\bm{a}$, measuring the quality of the mapping in terms of $U(\bm{s},\bm{a})$. Since the transmission rate is limited, the agent has to discard information in $\bm{s}$ that has little impact on $U$ to obtain a utility-maximizing $\bm{a}$ without exceeding the transmission limit $K$. Importantly, the constraint in transmission rate  directly translates into an instantaneous penalty signal leading to reduced utility, as outlined next for a one-step decision-making problem.

In a one-step scenario, we obtain the following
\begin{equation*}
\label{Eq:BoundedOne}
\max_{\pi_{\text{behave}}} \sum_{\bm{a} \in \mathcal{A}} \pi_{\text{behave}}(\bm{a}|\bm{s})U(\bm{s},\bm{a}) \ \ \text{s.t.} \ \ \text{KL}\left(\pi_{\text{behave}}||\pi_{\text{prior}}\right) \leq K, 
\end{equation*}
 where $\log \frac{\pi_{\text{behave}}(\bm{a}|\bm{s})}{\pi_{\text{prior}}(\bm{a}|\bm{s})}$ reflects instantaneous penalty\footnote{Note that although we use a state-independent prior in this work, the theoretical framework for Q-value reduction remains valid for state-conditioned $\pi_{\text{prior}}(\bm{a}|\bm{s})$.}. The above constrained optimization problem can be expressed as a concave unconstrained objective by introducing a Lagrange multiplier $\lambda > 0$:
\begin{equation}
\label{Eq:BoundedTwo}
\mathcal{L}^{\star}\left(\bm{s},\pi_{\text{prior}},\lambda\right)  = \max_{\pi_{\text{behave}}}  \sum_{\bm{a} \in \mathcal{A}} \pi_{\text{behave}}(\bm{a}|\bm{s})U(\bm{s},\bm{a}) 
-\frac{1}{\lambda} \text{KL}\left(\pi_{\text{behave}}||\pi_{\text{prior}}\right),
\end{equation}
where $\lambda$ trades off utility versus closeness to prior information. The optimum has a closed form:
\begin{equation}
\label{Eq:opt_pol}
\pi^{\star}_{\text{behave}}(\bm{a}|\bm{s}) = \frac{\pi_{\text{prior}}(\bm{a}|\bm{s})\exp\left(\lambda U(\bm{s},\bm{a})\right)}{\sum_{\bm{a}^{\prime}}\pi_{\text{prior}}(\bm{a}^{\prime}|\bm{s})\exp\left(\lambda U(\bm{s},\bm{a}^{\prime})\right)}. 
\end{equation}

Note that we are not the first to propose such information-theoretic principles within the context of RL (and planning), where the utility function is usually assumed to be the expected cumulative reward, i.e. $U(\bm{s}, \bm{a}) = Q(\bm{s}, \bm{a})$. In fact, similar principles have recently received increased attention within policy search and identification of optimal cumulative reward values, as outlined next.

In policy search, information-theoretic principles similar to Equation~\eqref{Eq:BoundedOne} can be categorized into three classes depending on the choice of the prior $\pi_{\text{prior}}(\bm{a}|\bm{s})$. The first class adopts a fixed prior that remains unchanged during learning. Entropy regularisation~\cite{Williams1991,Mnih2016} is a special case within this class (assuming a uniform prior policy). The second class uses a marginal prior policy obtained by averaging the behavioral policy over all environmental states. The information-theoretic intuition, here, is to encourage the agent to neglect reward-irrelevant information in the environment~\cite{Leibfried2015,Leibfried2016,Peng2017}. The third class assumes an adaptive prior (e.g. a policy learned at an earlier stage of training) to ensure incremental improvement steps in on-policy settings as learning proceeds~\cite{Bagnell2003,Peters2008,Peters2010,Schulman2015}. 

In optimal cumulative reward value identification, the KL-penalty is directly incorporated into Q-value estimates rather than using it for regularization. There are two distinct categories for value identification that utilize KL-constraints in different ways. The first category considers a restricted class of Markov Decision processes (MDPs), where instantaneous rewards incorporate a KL-penalty that explicitly discourages deviations from uncontrolled environmental dynamics. Such restricted MDPs enable efficient optimal value computation as outlined in~\cite{Todorov2009,Kappen2012}. The second category comprises MDPs with intrinsic penalty signals similar to Equation~\eqref{Eq:BoundedOne} where deviations from a prior policy are penalized. Optimal values are either computed with generalized value iteration schemes \cite{Tishby2011,Rubin2012,Grau-Moya2016}, or in an RL setting similar to Q-learning~\cite{Azar2012,Rawlik2012,Fox2016}.

Closest to our work are the recent approaches in~\cite{Haarnoja2017,Haarnoja2017b,Schulman2017}. It is worth mentioning that apart from the discrete action and high-dimensional state space setting, we tackle two additional problems not addressed previously. First, we consider \emph{dynamic} adaptation for trading off rewards versus intrinsic penalties as opposed to the static scheme presented in~\cite{Haarnoja2017,Haarnoja2017b,Schulman2017}. Second, we deploy a robust computational approach that incorporates value-based advantages to ensure bounded exponentiation terms. Our approach also fits into the work of how utilising entropy for reinforcement learning connects policy search to optimal cumulative reward value identification \cite{Haarnoja2017,Nachum2017,Donoghue2017,Schulman2017}. In this paper, however, we focus on deep value-based approaches, which show improved performance, as demonstrated in the experiments. 

Due to the intrinsic penalty signal, information-theoretic Q-learning algorithms provide a principled way of reducing Q-value estimates and are hence suited for addressing the overestimation problem outlined earlier. Although successful in the tabular setting, these algorithms are not readily applicable to high-dimensional environments that require parametric function approximators. In the next section, we adapt information-theoretic concepts to high-dimensional state spaces with function approximators and demonstrate that other deep learning techniques (e.g. DQNs) emerge as a special case. 

\subsection{Addressing Overestimation in Deep RL}
\label{sec:deep_info_rl}

We aim to reduce optimistic overestimation in deep RL methodologically by leveraging ideas from information theory. Since Q-value overestimations are a source of sample-inefficiency, we improve large-scale reinforcement learning where current techniques exhibit high sample complexity \cite{Mnih2015}.
To do so, we introduce an intrinsic penalty signal in line with the methodology put forward earlier. Before commencing, however, it can be interesting to gather more insights into the range of possible learners while tuning such a penalty. Plugging the optimal behavior policy $\pi^{\star}_{\text{behave}}$ from Equation~\eqref{Eq:opt_pol} back in Equation~\eqref{Eq:BoundedTwo} yields
\begin{equation*}
\label{Eq:FreeEnergy}
\mathcal{L}^{\star}(\bm{s},\pi_{\text{prior}},\lambda) = \frac{1}{\lambda} \log \sum_{\bm{a} \in \mathcal{A}}\pi_{\text{prior}}(\bm{a}|\bm{s})\exp\left(\lambda U(\bm{s},\bm{a})\right).
\end{equation*}

The Lagrange multiplier $\lambda$ steers the magnitude of the penalty and thus leads to different learning outcomes. If $\lambda$ is large, little penalization from the prior is introduced. As such, one would expect a learning outcome that mostly considers maximizing utility. This is confirmed as $\lambda \rightarrow \infty$, where
\begin{equation*}
\lim_{\lambda \rightarrow \infty} \mathcal{L}^{\star}(\bm{s},\pi_{\text{prior}}, \lambda) = \max_{\bm{a} \in \mathcal{A}} U(\bm{s},\bm{a}). 
\end{equation*}
On the other hand, for small $\lambda$ values, the deviation penalty is significant and the prior policy should dominate. This is again confirmed when $\lambda \rightarrow 0$, where we recover the expected utility under $\pi_{\text{prior}}$:
\begin{equation*}
\lim_{\lambda \rightarrow 0} \mathcal{L}^{\star}(\bm{s},\pi_{\text{prior}},\lambda) = \sum_{\bm{a} \in \mathcal{A}} \pi_{\text{prior}}(\bm{a}|\bm{s})U(\bm{s},\bm{a}) 
= \mathbb{E}_{\pi_{\text{prior}}}\left[U(\bm{s},\bm{a})\right].
\end{equation*}

Carrying this idea to deep RL by setting $U(\bm{s},\bm{a})=Q_{\bm{\theta}}(\bm{s},\bm{a})$, where $Q_{\bm{\theta}}(\bm{s},\bm{a})$ represents a deep Q-network, we notice that incorporating a penalty signal in the context of large-scale Q-learning with parameterized function approximators leads to 
\begin{equation*}
\mathcal{L}_{\bm{\theta}}^{\star}(\bm{s},\pi_{\text{prior}},{\lambda}) = \frac{1}{\lambda} \log \sum_{\bm{a} \in \mathcal{A}} \pi_{\text{prior}}(\bm{a}|\bm{s}) \exp\left(\lambda Q_{\bm{\theta}}(\bm{s},\bm{a})\right).
\end{equation*}
We use this operator to phrase an information-theoretic optimization objective for deep Q-learning: 
\begin{equation}
\label{Eq:Short}
\mathcal{J}_{\lambda}(\bm{\theta}) = \mathbb{E}_{\bm{s},\bm{a},r,\bm{s}^{\prime}} \left[\left(r + \gamma \mathcal{L}^{\star}_{\bm{\theta}^{-}}(\bm{s}^{\prime},\pi_{\text{prior}},{\lambda})  - Q_{\bm{\theta}}(
\bm{s},\bm{a})\right)^{2}\right],
\end{equation}
where $\mathbb{E}_{\bm{s},\bm{a},r,\bm{s}^{\prime}}$ refers to samples drawn from a replay memory in each iteration of training, and $\bm{\theta}^{-}$ to the parameter values at an earlier stage of learning. 

The above objective leads to a wide variety of learners and can be considered a generalization of current methods, including deep Q-networks~\cite{Mnih2015}. Namely, if $\lambda \rightarrow \infty$, we recover the approach in~\cite{Mnih2015} that poses the problem of optimistic overestimation:
\begin{equation*}
\mathcal{J}_{\lambda \rightarrow \infty} (\bm{\theta}) = 
\mathbb{E}_{\bm{s},\bm{a},r,\bm{s}^{\prime}}\Bigg[\Bigg(r + \gamma \max_{\bm{a}^{\prime} \in \mathcal{A}}Q_{\bm{\theta}^{-}}(\bm{s}^{\prime},\bm{a}^{\prime}) 
- Q_{\bm{\theta}}(\bm{s},\bm{a})\Bigg)^{2}
\Bigg]. 
\end{equation*}

On the contrary, if $\lambda \rightarrow 0$, we obtain the following
\begin{equation}
\label{Eq:ObjectiveLambda0}
\mathcal{J}_{\bm{\lambda} \rightarrow 0}(\bm{\theta}) = \mathbb{E}_{\bm{s},\bm{a},r,\bm{s}^{\prime}}\Bigg[\Bigg(r + \gamma \sum_{\bm{a}^{\prime}\in \mathcal{A}}\pi_{\text{prior}}(\bm{a}^{\prime}|\bm{s}^{\prime})Q_{\bm{\theta}^{-}}(\bm{s}^{\prime},\bm{a}^{\prime})
 - Q_{\bm{\theta}}(\bm{s},\bm{a})
\Bigg)^{2}\Bigg].
\end{equation}
Effectively, Equation~\eqref{Eq:ObjectiveLambda0} estimates future cumulative rewards using the prior policy as can be seen in the term $\sum_{\bm{a}^{\prime}\in \mathcal{A}}\pi_{\text{prior}}(\bm{a}^{\prime}|\bm{s}^{\prime})Q_{\bm{\theta}^{-}}(\bm{s}^{\prime},\bm{a}^{\prime})$. From the above two special cases, we recognize that our formulation allows for a variety of learners, where $\lambda$ steers outcomes between the above two limiting cases. Note, however, setting low values for $\lambda$ introduces instead a pessimistic bias~\cite{Fox2016}.
Since low $\lambda$-values introduce a pessimistic bias and large $\lambda$-values an optimistic bias, there must be a $\lambda$-value in between encouraging unbiased estimates. Unfortunately, it is not possible to compute such a $\lambda$ in closed form, which is why we propose a dynamical scheduling scheme based on temporal Bellman error evolution in the next section. Note that we assume a fixed prior $\pi_{\text{prior}}$ and we aim at scheduling $\lambda$. Another possibility would be to fix $\lambda$ and schedule the prior action probabilities instead. The latter is however practically less convenient compared to scheduling a scalar.

\section{Dynamic \& Robust Deep RL}
\label{sec:deep_info_rl_continued}

A fixed hyperparameter $\lambda$ is undesirable in the course of training as the effect of the intrinsic penalty remains unchanged. Since overestimations are more severe at the start of the learning process, a dynamic scheduling scheme for $\lambda$ with small values at the beginning (incurring strong penalization) and larger values towards the end (leading to less penalization) is preferable.

\paragraph{Adaptive $\lambda$:} 

A suitable candidate for dynamically adapting $\lambda$ in the course of training is the average squared loss (over replay memory samples) $\mathcal{J}_{\text{squared}} (t,p) = (t-p)^{2}$ between target values $t=r + \gamma \mathcal{L}^{\star}_{\bm{\theta}^{-}}(\bm{s}^{\prime},\pi_{\text{prior}}, \lambda)$ and predicted values $p=Q_{\bm{\theta}}(\bm{s},\bm{a})$.
The rationale, here, is that $\lambda$ should be inversely proportional to the average squared loss. If $\mathcal{J}_{\text{squared}}(t,p)$ is high on average, as is the case during early episodes of training, low values of $\lambda$ are favored. However, if $\mathcal{J}_{\text{squared}}(t,p)$ is low on average later in training, then high $\lambda$ values are more suitable for the learning process. 

We therefore propose to adapt $\lambda$ with a running average over the loss between targets and predictions. The running average $\mathcal{J}_{\text{avg}}$ should emphasize recent history as opposed to samples that lie further in the past since the parameters $\bm{\theta}$ of the Q-value approximator change over time. This is achieved with an exponential window and the online update
\begin{equation}
\label{eq:beta_updates}
\mathcal{J}_{\text{avg}} \leftarrow \left(1-\frac{1}{\tau}\right) \mathcal{J}_{\textrm{avg}} + \frac{1}{\tau} \mathbb{E}_{t,p} \left[ \mathcal{J}_{\text{squared}}(t,p) \right] ,
\end{equation}
where $\tau$ is a time constant referring to the window size of the running average, and $\mathbb{E}_{t,p} \left[ \mathcal{J}_{\text{squared}}(t,p) \right]$ is a shorthand notation for Equation~\eqref{Eq:Short}. This running average allows one to dynamically assign $\lambda = \frac{1}{\mathcal{J}_{\textrm{avg}}}$ at each training iteration.

The squared loss $\mathcal{J}_{\text{squared}}(t,p)$ has an impeding impact on the stability of deep Q-learning, where the parametric approximator is a deep neural net and parameters are updated with gradients and backpropagation. To prevent loss values from growing too large, the squared loss is practically replaced with an absolute loss if $|t-p|>1$~\cite{Mnih2015}, referred to as Huber loss $\mathcal{J}_{\text{Huber}}(t,p)$.
The Huber loss leads to a more robust adaptation of $\lambda$, as it uses an absolute loss for large errors instead of a squared one. Furthermore, the squared loss is more sensitive to outliers and might penalize the learning process unreasonably in the presence of sparse but large error values.

\paragraph{Robust Value Computation:}

The dynamic adaptation of $\lambda$ encourages learning of unbiased estimates of the optimal cumulative reward values. Presupposing $Q_{\bm{\theta}}(\bm{s},\bm{a})$ is bounded, $\mathcal{L}^{\star}_{\bm{\theta}}(\bm{s},\pi_{\text{prior}},\lambda)$ is also bounded in the limits of $\lambda$:
\begin{equation*}
\mathbb{E}_{\pi_{\text{prior}}}\left[Q_{\bm{\theta}}(\bm{s},\bm{a})\right] \leq \mathcal{L}^{\star}_{\bm{\theta}}(\bm{s},\pi_{\text{prior}},\lambda) \leq \max_{\bm{a} \in \mathcal{A}} Q_{\bm{\theta}}(\bm{s},\bm{a}). 
\end{equation*}

In practice, however, this operator is prone to computational instability for large $\lambda$ due to the exponential term $\exp\left(\lambda Q_{\bm{\theta}}(\bm{s},\bm{a})\right)$. We address this problem by amending the term $\frac{\exp\left(\lambda V_{\bm{\theta}}(\bm{s})\right)}{\exp\left(\lambda V_{\bm{\theta}}(\bm{s})\right)}$, where $V_{\bm{\theta}}(\bm{s}) = \max_{\bm{a}}Q_{\bm{\theta}}(\bm{s},\bm{a})$:
\begin{equation*}
\begin{split}
\mathcal{L}^{\star}_{\bm{\theta}} (\bm{s},\pi_{\text{prior}},\lambda) & = \frac{1}{\lambda} \log \sum_{\bm{a} \in \mathcal{A}} \pi_{\text{prior}}(\bm{a}|\bm{s}) \exp\left(\lambda Q_{\bm{\theta}}(\bm{s},\bm{a})\right) \frac{\exp\left(\lambda V_{\bm{\theta}}(\bm{s})\right)}{\exp\left(\lambda V_{\bm{\theta}}(\bm{s})\right)} \\
&=V_{\bm{\theta}}(\bm{s})+ \frac{1}{\lambda} \log \sum_{\bm{a} \in \mathcal{A}} \pi_{\text{prior}}(\bm{a}|\bm{s}) \exp\left(\lambda \left(Q_{\bm{\theta}}(\bm{s},\bm{a}) - V_{\bm{\theta}}(\bm{s})\right)\right)
\end{split}
\end{equation*}

The first term represents the maximum operator as in vanilla deep Q-learning. The second term is a log-partition sum with computationally stable elements due to the non-positive exponents $\lambda (Q_{\bm{\theta}}(\bm{s},\bm{a}) -V_{\bm{\theta}}(\bm{s})) \leq 0$. As a result, the log-partition sum is non-positive and subtracts a portion from $V_{\bm{\theta}}(\bm{s})$ that reflects cumulative reward penalization.

\begin{figure}[ht]
\vskip 0.2in
\begin{center}
\centerline{\includegraphics[width=0.7\columnwidth]{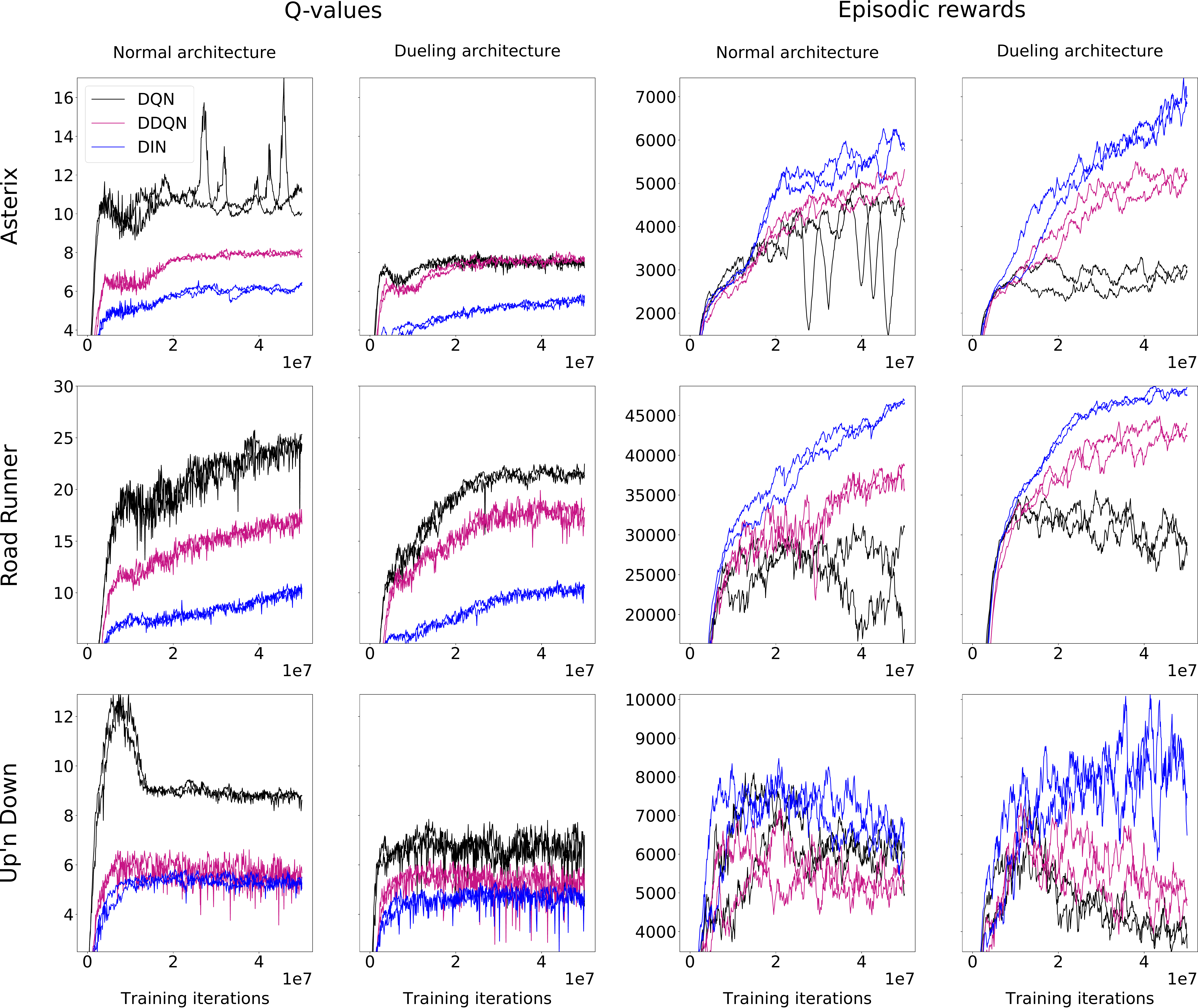}}
\caption{Q-values and episodic rewards for Asterix, Road Runner and Up'n Down for both normal and dueling architectures. 
Each plot shows three pairs of graphs, reporting the outcomes of two different random seeds, in black for DQN, purple for double DQN (DDQN) and blue for our information-theoretic approach (DIN). Clearly, our approach leads to lower Q-value estimates resulting in significantly better game play performance.}
\label{fig:3games}
\end{center}
\vskip -0.2in
\end{figure}

\begin{figure}[ht]
\vskip 0.2in
\begin{center}
\centerline{\includegraphics[width=0.4\columnwidth]{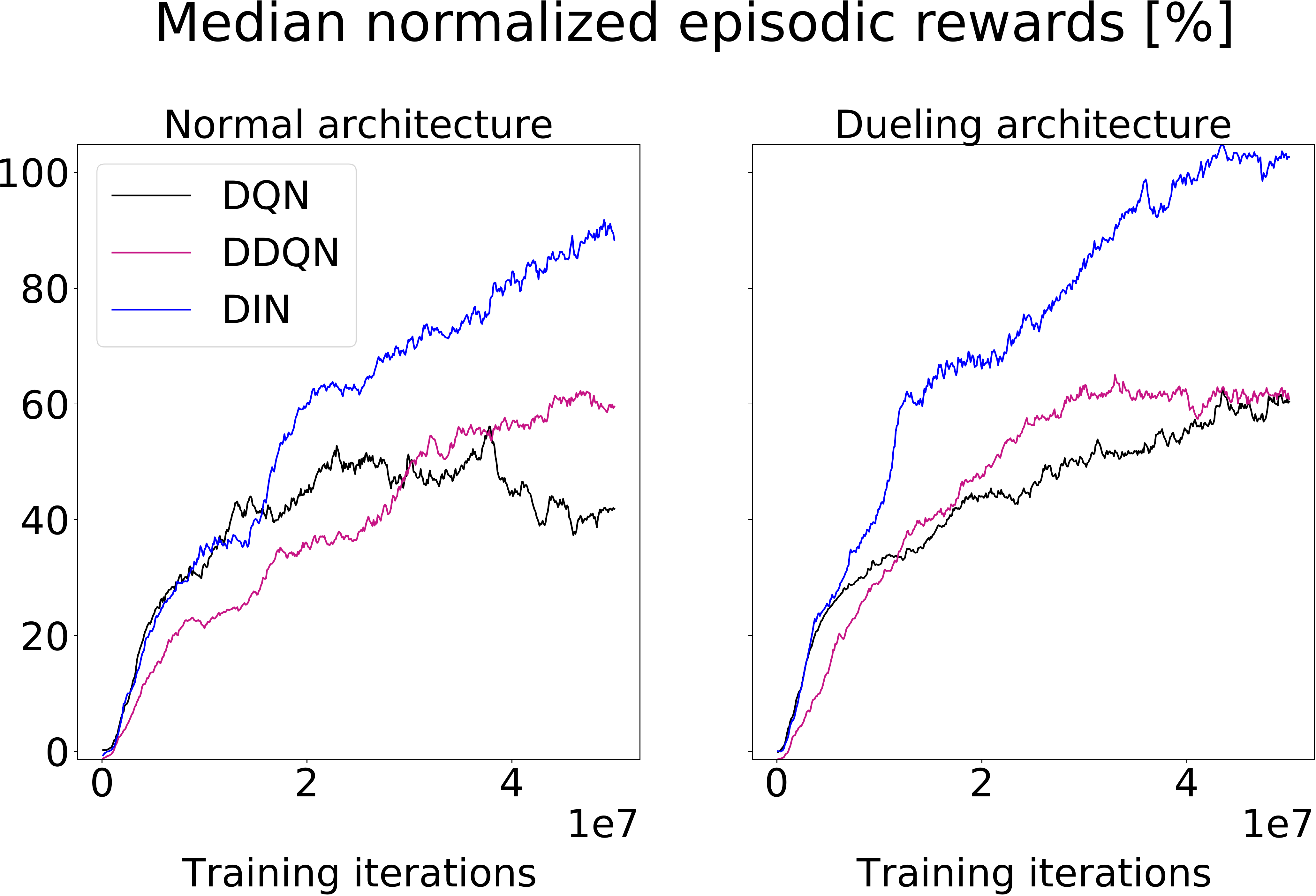}}
\caption{Median normalized episodic rewards across 20 Atari games for normal and dueling architectures. 
Each plot compares DQN (black), against double DQN (DDQN, purple) and our approach (DIN, blue). Our approach leads to significantly higher median game score.}
\label{fig:median_games}
\end{center}
\vskip -0.2in
\end{figure} 

\section{Experiments \& Results}

We hypothesize that addressing the overestimation problem results in improved sample efficiency and overall performance. To this end, we use the Atari domain \cite{Bellemare2013} as a benchmark to evaluate our method. We compare against deep Q-networks \cite{Mnih2015} that are susceptible to overestimations, and to double deep Q-networks \cite{vanHasselt2016}---an alternative proposed to address the precise problem we target. 
Our results demonstrate that our proposed method (titled deep information networks DIN) leads to significantly lower Q-value estimates resulting in improved sample efficiency and game play performance. We also show that these findings remain valid for the recently proposed dueling architecture~\cite{Wang2016}\footnote{Our approach could be incorporated into the newly released Rainbow framework~\cite{Hessel2018} that achieves state-of-the-art results by combining several independent DQN improvements over the past few years (one of them being double DQNs over which our approach achieves superior performance). Although we focus on Q-value identification in this work, ideas similar to DIN do apply as well to actor-critic methods like A3C~\cite{Mnih2016,Schulman2017}.}.

Parameter settings for reproducibility can be found in the appendix. We compare our approach against deep Q-networks and double deep Q-networks. We conduct further experiments by replacing network outputs with the dueling architecture \cite{Wang2016}. The dueling architecture leverages the advantage function $A(\bm{s},\bm{a})=Q(\bm{s},\bm{a})-\max_{\bm{a}}Q(\bm{s},\bm{a})$ and generalizes learning across actions. This results in improved game play performance, as confirmed in our experiments.

\subsection{Q-Values and Game Play Performance}
\label{q-values}

When training, networks are stored every $10^5$ iterations and used for offline evaluation. Evaluating a single network offline comprises $100$ game play episodes lasting for at most $4.5 \times 10^3$ iterations. In evaluation mode, the agent follows an $\epsilon$-greedy policy with $\epsilon=0.05$~\cite{Mnih2015}. We investigate 20 games.
Figure~\ref{fig:3games} reports results from the offline evaluation on three individual games (Asterix, Road Runner and Up'n Down), illustrating average maximum Q-values and average episodic rewards as a function of training iterations. Note that episodic rewards are smoothed with an exponential window, similar to Equation~\eqref{eq:beta_updates} with $\tau=10$, to preserve a clearer view. On all three games, our approach leads to significantly lower Q-value estimates when compared to DQN and double DQN for both, the normal and the dueling architecture (see left plots in Figure~\ref{fig:3games}). At the same time, this leads to significant improvements in game play performance (see right plots of Figure~\ref{fig:3games}).

Absolute episodic rewards ($\text{score}$) may vary substantially between different games. To ensure comparability across games, we normalize episodic rewards ($\text{score}_{\text{norm}}$) as $\text{score}_{\text{norm}} = \frac{\text{score} - \text{score}_{\text{random}}}{\text{score}_{\text{human}} - \text{score}_{\text{random}}} \cdot 100\% $,
where $\text{score}_{\text{random}}$ and $\text{score}_{\text{human}}$ refer to random and human baselines, see \cite{Mnih2015,Wang2016}.
Normalized episodic rewards enable a comparison across all 20 Atari games by taking the median normalized score over games \cite{Hessel2018}. The results of this analysis are depicted in Figure~\ref{fig:median_games} as a function of training iterations (smoothed with an exponential window using $\tau = 10$). Our approach clearly outperforms DQN and double DQN for both normal and dueling architectures. The dueling architecture yields an additional performance increase when combined with DIN. Our approach also yields superior results in terms of the best-performing agent (see the appendix for details). 

\subsection{Sample Efficiency}
\label{sampe_efficiency}

To quantify sample efficiency, we identify the minimal number of training iterations required to attain maximum deep Q-network performance. To this end, we compute the average episodic reward as in Figure~\ref{fig:3games} but smoothed with an exponential window $\tau=100$. We then identify for each approach the number of training iterations at which maximum deep Q-network performance is attained first. 

\begin{figure}[ht]
\vskip 0.2in
\begin{center}
\centerline{\includegraphics[width=0.8\columnwidth]{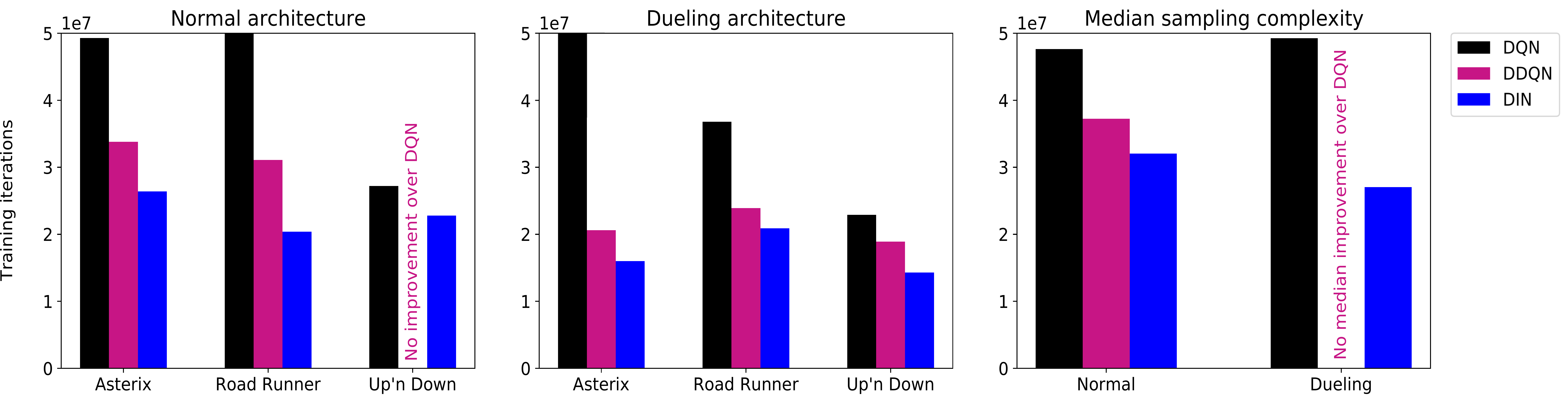}}
\caption{Sample efficiency for Asterix, Road Runner and Up'n Down under both normal and dueling architectures (left two panels) and when taking the median over 20 games (right panel). The color code is: DQN (black), double DQN (DDQN, purple) and our approach (DIN, blue). DINs are more sample-efficient for both architectures on the three games depicted and on average across 20 games.}
\label{fig:sample_eff}
\end{center}
\vskip -0.2in
\end{figure}

The results for Asterix, Road Runner and Up'n Down are shown in Figure~\ref{fig:sample_eff} in the left two panels. 
It can be seen that our approach leads to significant improvements in sample efficiency when compared to DQN and double DQN. For instance, DINs require only about $2 \times 10^7$ training iterations in Road Runner compared to about $3 \times 10^{7}$ for double DQNs, and about $5\times 10^{7}$ for standard DQNs using the normal architecture. These improvements are also valid for the dueling setting.
In order to assess sample efficiency across all 20 Atari games, we compute the median sampling efficiency over games, see Figure~\ref{fig:sample_eff} right panel. This analysis confirms the overall improved sample complexity attained in a wide range of tasks by our approach compared to DQN and double DQN.

\begin{figure*}[ht!]
\vskip 0.2in
\begin{center}
\centerline{\includegraphics[width=1\linewidth]{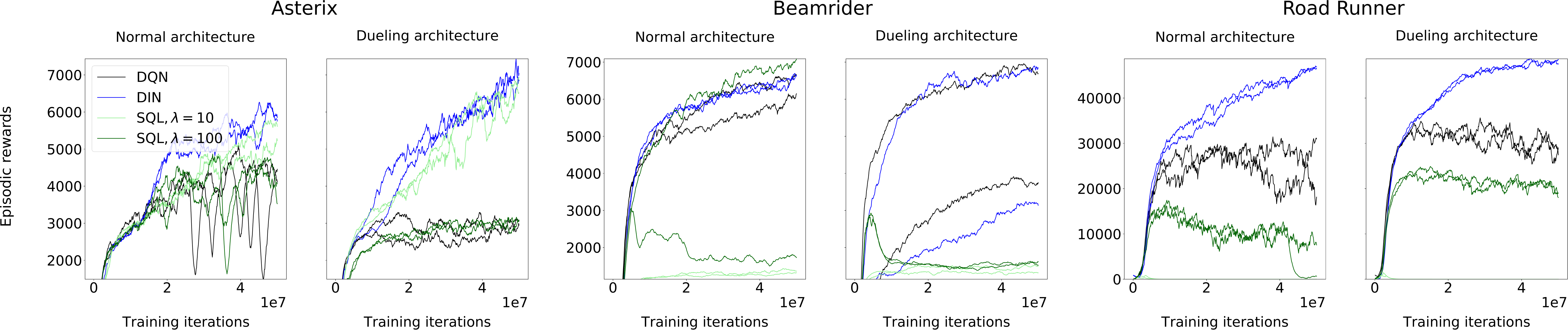}}
\caption{Episodic rewards for Asterix, Beamrider and Road Runner comparing our method to SQL. Clearly, our results show better performance in both the normal and dueling architecture without the necessity of identifying an optimal $\lambda$ in advance.}
\label{fig:q-values_and_score_rr_fixedLambda}
\end{center}
\vskip -0.2in
\end{figure*}

\subsection{Comparison to Soft Q-Learning (SQL)}

The closest work to our approach is that of~\cite{Schulman2017}, where the authors consider information theory to bridge the gap between Q-learning and policy gradients RL. Our approach goes further by considering dynamic adaptation for $\lambda$ in the course of training, and introduces robust computation based on value advantages. We compare our method to SQL (where $\lambda$ is fixed) on the games Asterix, Beamrider and Up'n Down. Results depicted in Figure~\ref{fig:q-values_and_score_rr_fixedLambda} demonstrate that our method can outperform SQL on these three games by significant margins without the requirement of pre-specifying $\lambda$. For instance, DINs achieve the best performance of SQL in about 5,000,000 iterations on the Road Runner game.

\section{Conclusions}

In this paper, we proposed a novel method for reducing sample complexity in deep reinforcement learning. Our technique introduces an intrinsic penalty signal by adapting principles from information theory to high-dimensional state spaces. We showed that DQNs are a special case of our proposed approach for a specific choice of the Lagrange multiplier steering the intrinsic penalty. Finally, in a set of experiments on 20 Atari games, we demonstrated that our technique indeed outperforms competing approaches in terms of performance and sample efficiency. These results remain valid for the dueling architecture from \cite{Wang2016} yielding a further performance boost.

% The most promising direction of future work is to study adaptive prior policies instead of fixed ones, in line with \cite{Bagnell2003,Peters2008,Peters2010,Schulman2015}. This could be used to extend our framework to multi-task learning scenarios where task-specific policies satisfy a KL-constraint to prevent deviation from a common prior. The common prior can encode a behavioral policy that generalizes across tasks, thus enabling knowledge transfer between problems with a shared latent structure. 

\bibliography{icml_paper}
\bibliographystyle{unsrt}

\appendix

\section{Training Details}
We conduct all experiments in Python with TensorFlow and OpenAI gym extending the GitHub project from~\cite{Kim2016}. We use a deep convolutional neural network $Q_{\bm{\theta}}(\bm{s},\bm{a})$ as a function approximator for Q-values, designed and trained according to~\cite{Mnih2015}. $Q_{\bm{\theta}}(\bm{s},\bm{a})$ receives as input the current state of the environment $\bm{s}$ that is composed of the last four video frames. The number of neurons in the output layer is set to be the number of possible actions $\bm{a}$. Numerical values of each output neuron correspond to the expected cumulative reward when taking the relevant action in state $\bm{s}$.

We train the network for $5 \times 10^7$ iterations where one iteration corresponds to a single interaction with the environment. Environment interactions $(\bm{s},\bm{a},r,\bm{s}^{\prime})$ are stored in a replay memory consisting of at most $10^6$ elements. Every fourth iteration, a minibatch of size $32$ is sampled from the replay memory and a gradient update is conducted with a discount factor $\gamma=0.99$. We use RMSProp \cite{Tieleman2012} as the optimizer with learning rate $2.5 \times 10^{-4}$, gradient momentum $0.95$, squared gradient momentum $0.95$, and minimum squared gradient $0.01$. Rewards $r$ are clipped to $\{-1,0,1\}$. The target network $Q_{\bm{\theta}^{-}}(s,a)$ is updated every $10^{4}$ iterations. The time constant $\tau$ for dynamically updating the hyperparameter $\lambda$ is $10^5$, and the prior policy $\pi_{\text{prior}}$ is uniform. A uniform prior ensures a pessimistic baseline in case of small $\lambda$. This pessimistic baseline guarantees the existence of unbiased $\lambda$-configurations our scheduling scheme aims to detect.

When the agent interacts with the environment, every fourth frame is skipped and the current action is repeated on the skipped frames. During training, the agent follows an $\epsilon$-greedy policy where $\epsilon$ is initialized to $1$ and linearly annealed over $10^6$ iterations until a final value of $\epsilon = 0.1$. Training and $\epsilon$-annealing start at $5 \times 10^4$ iterations. RGB-images from the Arcade Learning Environment are preprocessed by taking the pixel-wise maximum with the previous image. After preprocessing, images are transformed to grey scale and down-sampled to $84 \times 84$ pixels. All our experiments are conducted in duplicate with two different initial random seeds. The random number of NOOP-actions at the beginning of each game episode is between $1$ and $30$.

\section{Policy Evaluation}
\label{policy}

We compare the performance of all approaches in terms of the best (non-smoothed) episodic reward (averaged over 100 episodes) obtained in the course of the entire evaluation procedure. To ensure comparability between games, we again make use of normalized scores described earlier.

Our results are summarized for the normal and dueling architecture in Tables~\ref{tab:overall_res} and~\ref{tab:overall_res_duel} respectively. In both cases, our approach achieves superior median normalized game performance compared to DQN and double DQN. In the normal setting, DIN achieves best performance across all three approaches in 11 out of 20 games, whereas in the dueling setting, DIN achieves best performance in 13 out of 20 games. We can confirm that the dueling architecture, when combined with DIN, leads to a performance increase in 15 out of 20 games, which is not reflected in the median performance.

\begin{table}[ht]
\caption{Normalized episodic rewards (normal architecture).}
\label{tab:overall_res}
\vskip 0.15in
\begin{center}
\begin{small}
\begin{sc}
\begin{tabular}{ |c||c|c|c| }
 \hline
 Game & DQN & DDQN & DIN\\ 
 \hline
 \hline
 Assault & 198.8\% & 214.9\% & \textbf{233.5\%} \\
 Asterix & 70.4\% & 73.4\% & \textbf{85.0\%} \\
 Bank Heist & 76.2\% & 68.3\% & \textbf{87.8\%} \\
 Beamrider & 123.6\% & 117.5\% & \textbf{127.2\%} \\
 Berzerk & \textbf{35.8\%} & 33.7\% & 22.3\% \\
 Double Dunk & 161.6\% & \textbf{265.8\%} & 165.8\% \\
 Fishing Derby & 205.4\% & \textbf{211.2\%} & 202.4\% \\
 Freeway & \textbf{103.3\%} & 75.2\% & 101.6\% \\
 Kangaroo & 129.8\% & 94.4\% & \textbf{137.1\%} \\
 Krull & 401.9\% & 494.8\% & \textbf{534.6\%} \\
 Kung Fu Master & 130.0\% & -0.8\% & \textbf{144.5\%} \\
 Qbert & 22.1\% & \textbf{31.9\%} & 21.4\% \\
 Riverraid & 14.1\% & 20.4\% & \textbf{26.1\%} \\
 Road Runner & 503.6\% & 593.2\% & \textbf{643.7\%} \\
 Seaquest & \textbf{4.0\%} & 1.2\% & 2.9\% \\
 Space Invaders & 53.9\% & 51.1\% & \textbf{54.0\%} \\
 Star Gunner & 560.5\% & 571.1\% & \textbf{595.0\%} \\
 Time Pilot & 40.1\% & \textbf{175.0\%} & 171.4\% \\
 Up'n Down & 131.5\% & \textbf{135.8\%} & 135.2\% \\
 Video Pinball & 4385.8\% & \textbf{5436.6\%} & 4654.1\% \\
 \hline
 \hline
 Median & 126.7\% & 106.0\% & \textbf{136.2\%} \\
 \hline
\end{tabular}
\end{sc}
\end{small}
\end{center}
\vskip -0.1in
\end{table}

\begin{table}[ht]
\caption{Normalized episodic rewards (dueling architecture).}
\label{tab:overall_res_duel}
\vskip 0.15in
\begin{center}
\begin{small}
\begin{sc}
\begin{tabular}{ |c||c|c|c| }
 \hline
 Game & DQN & DDQN & DIN\\ 
 \hline
 \hline
 Assault & 260.4\% & 269.5\% & \textbf{336.6\%} \\
 Asterix & 45.6\% & 75.7\% & \textbf{104.1\%} \\
 Bank Heist & 74.2\% & 78.5\% & \textbf{79.1\%} \\
 Beamrider & \textbf{130.6\%} & 128.4\% & 129.3\% \\
 Berzerk & 32.5\% & \textbf{33.4\%} & 24.6\% \\
 Double Dunk & 223.9\% & \textbf{241.6\%} & 222.6\% \\
 Fishing Derby & \textbf{177.2\%} & 163.4\% & 29.3\% \\
 Freeway & 103.8\% & 102.9\% & \textbf{104.6\%} \\
 Kangaroo & 347.6\% & 186.8\% & \textbf{437.4\%} \\
 Krull & 480.8\% & 433.6\% & \textbf{574.3\%} \\
 Kung Fu Master & 114.8\% & 118.4\% & \textbf{129.6\%} \\
 Qbert & 70.2\% & \textbf{84.2\%} & 82.8\% \\
 Riverraid & \textbf{28.4\%} & 22.9\% & 22.8\% \\
 Road Runner & 553.6\% & 624.4\% & \textbf{659.0\%} \\
 Seaquest & \textbf{14.4\%} & 0.5\% & 5.1\% \\ 
 Space Invaders & 63.1\% & 26.7\% & \textbf{140.6\%} \\
 Star Gunner & 139.5\% & 145.1\% & \textbf{169.4\%} \\
 Time Pilot & 109.6\% & 56.7\% & \textbf{265.1\%} \\
 Up'n Down & 105.5\% & 125.6\% & \textbf{150.9\%} \\  
 Video Pinball & 4461.6\% & 4754.5\% & \textbf{4982.8\%} \\
 \hline
 \hline
 Median & 112.2\% & 122.0\% & \textbf{135.1\%} \\
 \hline
\end{tabular}
\end{sc}
\end{small}
\end{center}
\vskip -0.1in
\end{table}

\end{document}